%% file: incremental_configuration_preprint.tex
\documentclass{ecai}
\usepackage{times}
\usepackage{graphicx}
\usepackage{latexsym}
\usepackage{url}
\usepackage{csquotes}
\usepackage{amsmath}
\usepackage{amssymb}
\usepackage{xspace}
\usepackage{listings}
\usepackage{cite}

\ecaisubmission   

\usepackage[capitalize]{cleveref}

\input{macros}

\begin{document}


\title{Applying Incremental Answer Set Solving\\to Product Configuration\footnote{This is the authors' version of the work. It is posted here for your personal use. Not for redistribution. The definitive version is published as https://doi.org/10.1145/3503229.3547069.}}

\author{
	Richard Comploi-Taupe \and
	Giulia Francescutto \and
	Gottfried Schenner \institute{
		Siemens, Austria,
		email: richard.taupe@siemens.com
	}
}

\maketitle
\bibliographystyle{ecai}

\begin{abstract}
	In this paper, we apply incremental answer set solving to product configuration.
	Incremental answer set solving is a step-wise incremental approach to Answer Set Programming (ASP).
	We demonstrate how to use this technique to solve product configurations problems incrementally.
	Every step of the incremental solving process corresponds to a predefined configuration action.
	Using complex domain-specific configuration actions makes it possible to tightly control the level of non-determinism and performance of the solving process.
	We show applications of this technique for reasoning about product configuration, like simulating the behavior of a deterministic configuration algorithm and describing user actions.

\end{abstract}

\section{INTRODUCTION}
In many industrial environments, complex customizable products and services need to be configured by selecting and validating hundreds, if not thousands, of features. An efficient configuration tool is crucial for reaching the goal of performing this process in an acceptable time. However, the more features are involved, the more difficult this task becomes.

Declarative programming paradigms, like Answer Set Programming (ASP) \cite{DBLP:books/sp/Lifschitz19} or Constraint Programming \cite{rossi2006handbook}, are established methods for solving product configuration problems \cite{soininen2001representing,freuder1998role,fleischanderl1998configuring}. Although domain-specific deterministic configuration algorithms can outperform purely declarative solving approaches, especially in large-scale configuration problems, declarative paradigms are still preferable for several reasons. 

In case of domain-specific algorithms implemented in some imperative programming language, it is hard to verify if the algorithm works correctly for every given input. Additionally, if a new constraint is added to the domain, the algorithm must be adapted to consider the new constraint without violating any existing constraint. In a declarative system this problem does not exist, as the separation of declarative knowledge and solving guarantees the correctness of the result.

Another important aspect particularly relevant for engineering configurators in industry are complex user actions \cite{felfernig2010personalized,DBLP:conf/confws/FalknerHKST19,DBLP:journals/jucs/FalknerHKSST20}.
Whereas in simple consumer configurators a user typically selects attribute values, e.g., \enquote{choose color of t-shirt,\!} in an engineering configurator a user action can lead to the creation of hundreds of sub-components, e.g., \enquote{configure power supply for railway system.\!} 
The challenge of creating a user interface in such a system is to define, on one side, complex user actions that are intuitive for a domain expert, and on the other side, to make the configuration process as efficient as possible, i.e., allow to configure the system in a small number of steps \cite{rabiser2012qualitative}.

In this work, we use ASP to tackle a typical hardware configuration problem. We take advantage of the multi-shot \cite{gebser_kaminski_kaufmann_schaub_2019} capabilities introduced by \slv{clingo} with the intention to alleviate the grounding bottleneck that arises when the configuration size is not known and a large enough upper bound on the domain size needs to be defined.

Multi-shot solving, in contrast to the usual single-shot ASP, supports splitting the ASP program into sub-programs that can be grounded separately, giving the possibility to approach an extremely complex problem by grounding parts of it step-wise until a certain goal is reached. Incremental ASP is a special form of multi-shot reasoning where the program has a base part that represents static knowledge, and a step part, that captures knowledge that evolves with increasing step $t$. 

Using incremental Answer Set Programming we discuss various encodings to solve a typical product configuration problem. We will make use of this functionality to simulate a deterministic configuration algorithm and describe complex user actions using incremental ASP solving.

The paper is structured as follows: in \cref{sec:preliminaries} we give some ASP preliminaries, then we introduce the incremental approach and the example domain in \cref{sec:solving}. In \cref{sec:encodings}, the different ASP encodings with incremental solving for configuration problems are shown, and the experimental results are presented in \cref{sec:experiments}.

\section{PRELIMINARIES}
\label{sec:preliminaries}
In this section, we present preliminaries on (multi-shot) Answer Set Programming.

\subsection{Answer Set Programming}
As usual, a normal logic program is a set of rules $r$ of the form
\begin{equation}
	h \gets b_1, \ldots, b_m, \mathrm{not}\; b_{m+1}, \ldots, \mathrm{not}\; b_n.
\end{equation}
where $h$ and $b_1,\ldots,b_n$ are atoms, with $n\ge 0$. Each \emph{atom} is an expression of the form $p(t_1,\ldots,t_k)$, where $p$ is a predicate symbol and $t_i$, for $i=1,\ldots,k$ are \emph{terms}. Terms are composed of either constants, variables or function symbols. The symbol $\mathit{not}$ stands for \emph{negation as failure}.

We denote with $H(r) = h$ the \emph{head} of a rule $r$, with $B(r) = \{ b_1, \ldots, b_m, \mathrm{not}\; b_{m+1}, \ldots, \mathrm{not}\; b_n \}$ the \emph{body}, with $B^+(r) = \{b_1, \ldots, b_m \}$ the positive body and with $B^-(r) = \{ b_{m+1}, \ldots, b_n \}$ the negative body of a rule. A rule $r$ is called a \emph{fact} when $B(r) = \emptyset$ and a \emph{constraint} when $H(r) = \bot$.

A term without variables occurrences is said to be \emph{ground}; a \emph{ground} instance $G_P$ of a program $P$ is obtained by substituting the variables in each rule $r$ of $P$ with ground terms appearing in $P$.
A set of ground atoms $M$, with $\bot \notin M$, satisfies a rule $r \in G_P$ if $B^+(r) \subseteq M$ and $B^-(r) \cap M = \emptyset$ imply $H(r)\in M$. $M$ is a \emph{model} of $P$ if it satisfies each $r\in G_P$.

$M$ is called a \emph{stable} model of $P$ \cite{DBLP:conf/iclp/GelfondL88}, or \emph{answer set}, if it is a $\subseteq$-minimal model of the reduct 
$\{H(r) \gets B^+(r) \mid r \in G_P, B^-(r) \cap M =
\emptyset\}$. 

\subsection{Multi-shot solving in ASP}
\slv{clingo} provides an extension to the classic ASP language that enables the use of the \emph{multi-shot} solving paradigm \cite{gebser_kaminski_kaufmann_schaub_2019}. This extension allows for splitting an ASP program into subparts, and to interleave grounding and solving of such parts by exercising control via an imperative programming interface.

To achieve this, new \#{}program directives are introduced in the ASP language. Each subprogram declaration is of form 
\begin{lstlisting}[mathescape=true]
	#program $n(p_1,\ldots,p_k)$
\end{lstlisting} 
where $n$ is the name of the subprogram, and $p_1,\ldots,p_k$ are its parameters. Each subprogram gathers all the rules up to the next subprogram declaration. All the rules not preceded by any \#{}program declaration belong to a dedicated \#{}program base.

Control over ground and solve steps is exercised via an imperative programming \slv{clingo} API, where the internal state of \slv{clingo} is represented by a control object. \#{}external directives are used to set external atoms to some truth value via the \slv{clingo} API.

\section{SOLVING PRODUCT CONFIGURATION PROBLEMS WITH INCREMENTAL ASP}
\label{sec:solving}
In this section, we present our approach to solving configuration problems with incremental, multi-shot answer set solving. In particular, we will apply incremental Answer Set Solving to the \enquote{Hardware Racks} problem.

\subsection{ Example domain \enquote{Hardware Racks} }

\begin{figure}
	\includegraphics[width=\linewidth]{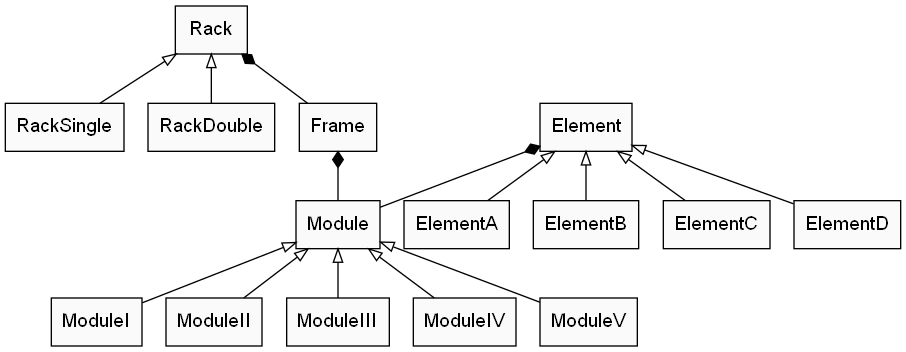}
	\caption{Hardware Racks Domain}
	\label{fig:racks_uml}
\end{figure}

The example domain describes a typical hardware configuration problem found in industry.
Its UML class diagram is shown in \cref{fig:racks_uml}.
The example domain is a variation of the running example in \cite{falkner2015ooasp}, where a generic framework for solving product configuration problems in a non-incremental manner has been described.

The example domain \enquote{hardware racks} contains these concepts:
\begin{itemize}
	\item Racks of type rackSingle, rackDouble
	\item Frames
	\item Elements of type elementA, elementB, elementC, elementD
	\item Modules of type moduleI, moduleII, moduleIII, moduleIV, moduleV
	\item Racks contain Frames
	\item Frames contain Modules
	\item Elements require Modules
\end{itemize}

Additionally, these constraints hold:

\begin{itemize}
	\item A RackSingle contains exactly 4 Frames
	\item A RackDouble contains exactly 8 Frames
	\item A Frame can contain up to 5 Modules
    \item A Frame is contained in exactly 1 Rack
	\item A Module is contained in exactly 1 Frame
 	\item A Module requires 0/1 Elements
	\item An elementA/B/C/D requires 1/2/3/4 moduleI/II/III/IV instances
	\item A frame can contain a moduleII, iff it also contains exactly one moduleV
\end{itemize}

A valid configuration is an instantiation of the domain model in which no domain constraints are violated.

\subsection{Representing configuration problems for incremental ASP}

In the incremental solving approach, configurations are constructed in a step-wise manner.

In our encoding the configuration at step STEP is described with atoms of the form $\mathrm{configuration}(\mi{FACT}, \mi{STEP})$.

The facts describe the instances, attributes and association values in the current domain.
For the hardware rack domain the following FACTs are available:
\begin{itemize}
	\item $\mathrm{isA}(\mi{OBJID},\mi{CLASSNAME})$
 	\item $\mathrm{element\_module}(\mi{ID1},\mi{ID2})$
 	\item $\mathrm{rack\_frame}(\mi{ID1},\mi{ID2})$
  	\item $\mathrm{frame\_module}(\mi{ID1},\mi{ID2})$
\end{itemize}
Every object has a unique object id (within the configuration) and must be the instance of exactly one leaf class.

To describe the constraints of the domain two different mechanisms are used.
Upper bound constraints are described with ASP constraints and can therefore never be violated while constructing a solution.
For example, in Listing \ref{prg:inc:upperbound}, the constraint guarantees that a RackSingle will never have more than 4 Frames.

\begin{lstlisting}[mathescape=true,escapeinside={\#(}{\#)},basicstyle={\ttfamily\small},label=prg:inc:upperbound,caption={Upper bound constraint},linerange={138-139}]
:- N>4, configuration(isA(R,rackSingle),t), 
   rack_nrofframes(R,N,t). 
\end{lstlisting}

Lower bound constraints are constraints that may be violated in the current configuration,
but can become satisfied at a later stage of the configuration process.
For this kind of constraints, $\mathrm{cv}(\mi{DESCRIPTION},\mi{STEP})$ atoms are used (see Listing \ref{prg:inc:lowerbound}) to describe constraint violations (cv) of the partial configuration in the current step.

\begin{lstlisting}[mathescape=true,escapeinside={\#(}{\#)},basicstyle={\ttfamily\small},label=prg:inc:lowerbound,caption={Lower bound constraint},linerange={134-135}]
cv(rack_needs_more_frames(R),t) :- 
   N<4, rack_nrofframes(R,N,t). 
\end{lstlisting}

\subsection{Incrementally solving a configuration problem}

All the discussed encodings use the same solving approach. The idea is to build the configuration step by step, and at each step, a check is done to see if there are still constraints violated, hence the configuration is not yet completed.

Initially a configuration containing a fixed set of elements is given. For instance, in Listing \ref{prg:inc:initial} the initial configuration includes one elementA with OBJID $1$.
\begin{lstlisting}[mathescape=true,escapeinside={\#(}{\#)},basicstyle={\ttfamily\small},label=prg:inc:initial,caption={Initial configuration at step 0},linerange={24-24}]
configuration(isA(1,elementA),0). 
\end{lstlisting}

Then, at each step of the incremental solving, a set of possible actions is determined based on the configuration of the previous step. In Listing \ref{prg:inc:possibleaction}, one possible action at step $t$ is the creation of a frame for rack $R$, since in step $t-1$, rack $R$ did not have enough frames, and this caused the violated constraint \lstinline{cv(racks_needs_more_frames(R),t-1)}.
\begin{lstlisting}[mathescape=true,escapeinside={\#(}{\#)},basicstyle={\ttfamily\small},label=prg:inc:possibleaction,caption={Possible action at step $t$},linerange={96-97}]
possible(action(create_frames_for(R)),t) :- 
   cv(rack_needs_more_frames(R),t-1). 
\end{lstlisting}

From the set of possible actions at step $t$, exactly one action is chosen and executed (Listing \ref{prg:inc:actionchoice}).
\begin{lstlisting}[mathescape=true,escapeinside={\#(}{\#)},basicstyle={\ttfamily\small},label=prg:inc:actionchoice,caption={Action choice at step $t$},linerange={44-45}]
1 { do(X,t):possible(X,t) } 1 :- 
   action_possible(t). 
\end{lstlisting}

An action adds one or more facts to the configuration.
This process is continued until a valid configuration is found, i.e., no constraints are violated (Listing \ref{prg:inc:query}).
\begin{lstlisting}[mathescape=true,escapeinside={\#(}{\#)},basicstyle={\ttfamily\small},label=prg:inc:query,caption={Iteration guard}, linerange={30-30}]
:- cv(CV,t), query(t). 
\end{lstlisting}

\section{INCREMENTAL ENCODINGS}
\label{sec:encodings}

Depending on the choice of possible actions, level of non-determinism and the granularity of the actions, many different strategies for solving configuration problems are possible.
In the following we describe different encodings, each representing a typical solving approach.

\subsection{ Generic Encoding }
\label{sec:encodings:generic}

The generic encoding uses a small set of generic actions.
It is suitable for any domain because it doesn't use any domain-specific actions.
In the generic encoding, the following actions are possible:

\begin{itemize}
	\item $\mathrm{create\_object}(C)$: \mi{C} must be a leaf class
    \item $\mathrm{associate}(\mi{ASSOC},\mi{ID1},\mi{ID2})$: \mi{ID1} and \mi{ID2} are IDs of existing objects
\end{itemize}

\begin{figure}
	\includegraphics[width=\linewidth]{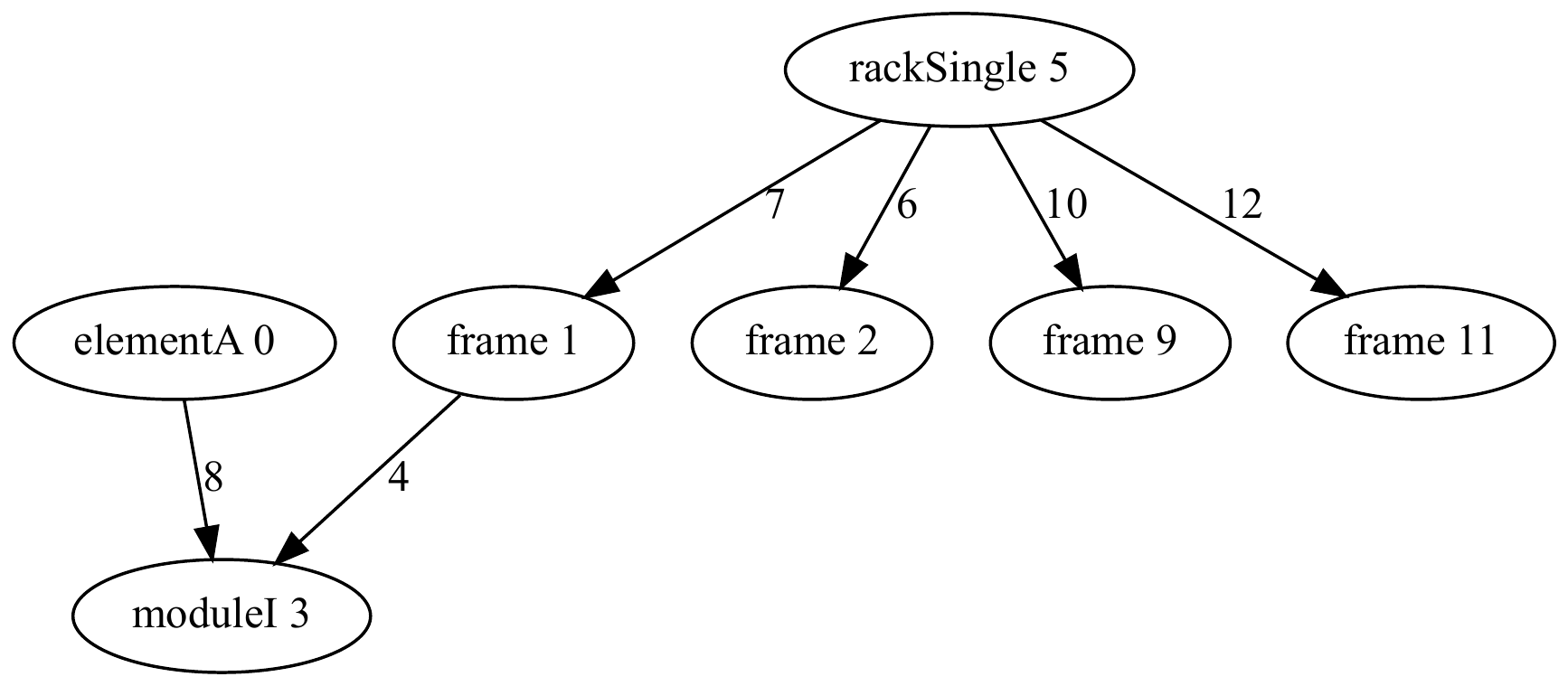}
	\caption{Generic solving example}
	\label{fig:racks_generic_example}
\end{figure}

In the generic encoding, at each step either a new object is created or some existing objects are associated. 
\Cref{fig:racks_generic_example} shows a configuration constructed by the generic encoding.
The numbers indicate the step at which the part of the configuration (object or association link) was added to the configuration.
As can be seen there is no particular order in which the actions were taken. 
One ElementA was given (step 0), then two frames (step 1 and 2) and a module (step 3) was created.
In step 4, the frame created in step 1 and the module were associated. 
The final configuration took 12 steps, which is the maximum number of required solving steps of all encodings.

In a way, the generic encoding is the most flexible encoding as it is capable of creating every possible configuration.
If new constraints are added to the domain there is no need to change the actions of the encoding.
On the downside, because of the fine granularity of the actions, the encoding will always need the maximum number of steps to construct a solution.
Beside that, it will also construct identical solutions in different ways because of the lack of symmetry breaking constraints and domain-specific knowledge.

\subsection{ Encoding with predefined order of action }
\label{sec:encodings:ordered}

The ordered encoding uses a specific order of domain-specific actions to optimize the solving process.
The possible actions in order are:

\begin{enumerate}
	\item $\mathrm{create\_modules\_for\_element}(E)$: create all required modules for an element
 	\item $\mathrm{create\_frame\_for\_module}(M)$: create/assign frame for module
 	\item $\mathrm{create\_rack\_for\_frame}(F)$: create/assign rack to frame
    \item $\mathrm{create\_frame\_for\_rack}(R)$: create frames for rack
\end{enumerate}

The actions are ordered, i.e., an action of a higher level can only be taken, if no lower level action is applicable.
For example, $\mathrm{create\_frame\_for\_module}$ can only be applied if all $\mathrm{create\_modules\_for\_element}$ have been executed.
Also, in contrast to the generic encoding, one action can create multiple objects and associate multiple objects.
For instance, the action for creating the modules of an ElementD creates in one step four instances of ModuleIV and associates it with the element.

\begin{figure}
	\includegraphics[width=\linewidth]{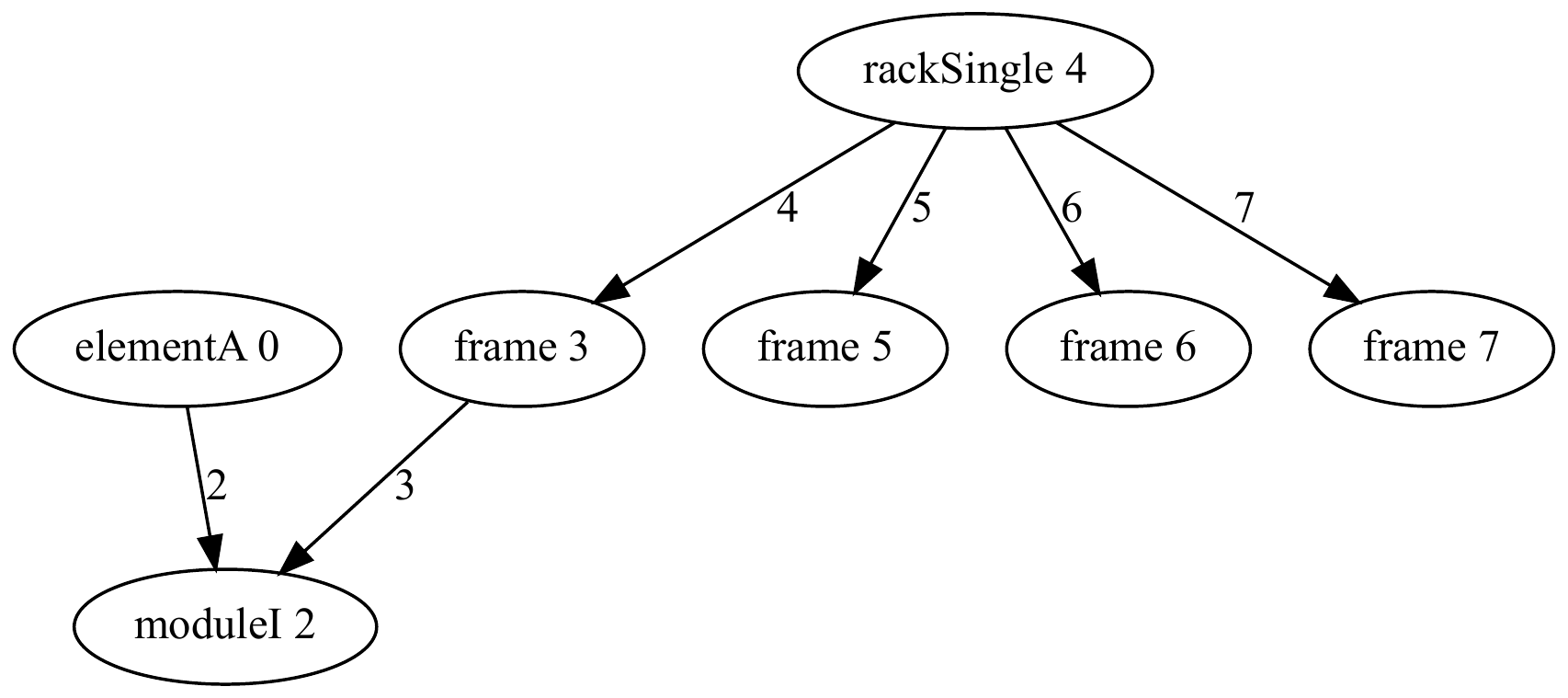}
	\caption{Ordered solving example}
	\label{fig:racks_ordered_example}
\end{figure}

\Cref{fig:racks_ordered_example} shows the solving process for the same input as for \cref{fig:racks_generic_example}.
Now, the configuration is constructed in fewer steps and in a particular order.
First the module is created for the element, then a frame for the module, then the rack and finally the missing frames of the rack.
This corresponds to the defined order of the actions.

The actions of this encoding are domain-specific, e.g the creation of modules for an element.
If a new constraint is added to the domain, not only the constraint must be formalized but it might also be necessary to change the domain-specific actions.
In that sense this encoding is less declarative than the generic encoding.

\subsection{ Algorithmic encoding }
\label{sec:encodings:algorithm}

The algorithmic encoding is a deterministic version of the ordered encoding.
It uses the same actions, but the action at every step is chosen deterministically.
If multiple actions are possible, the solver chooses the first one available.
Therefore the encoding must ensure that possible actions lead to valid solutions.

Although for the simple example the constructed configuration is the same as in \cref{fig:racks_ordered_example}, the algorithmic approach constructs the configuration deterministically.

The difference between the deterministic and the non-deterministic version can be seen from the example of the $\mathrm{create\_rack\_for\_frame}(F)$ action.
In the non-deterministic case all existing racks (and a potential new rack) that do not violate an upper bound can be part of a (alternative) solution.
In the deterministic case only one suitable rack must be determined with ASP rules (see Listing \ref{prg:alg:usable}).
If a $\mathrm{first\_usable\_rack}$ can be derived for the current configuration, the frame will assigned to this rack, otherwise a new rack will be created.
\begin{lstlisting}[mathescape=true,escapeinside={\#(}{\#)},basicstyle={\ttfamily\small},label=prg:alg:usable,caption={Usable racks for deterministic case}, linerange={74-79}]
usable_rack(R,F,t) :- 
   configuration(isA(F,frame),t), 
   rack_nrofframes(R,N,t), N<4 . 
first_usable_rack(R,F,t) :- 
   usable_rack(R,F,t), 
   { usable_rack(ROTHER,F,t): ROTHER<R } 0.
\end{lstlisting}

\paragraph{Application.}

The algorithmic encoding was inspired by domain-specific hardware configuration algorithms implemented in an imperative programming language.
As these algorithms are deterministic there is usually no other way than testing to ensure that the algorithms are correct.
By expressing the algorithm in ASP, it is now possible to reason if the algorithm violates certain constraints.

For example, suppose that there is a new constraint, that all modules of an element must be in the same frame.
Testing the configuration algorithm by running it for small examples might give the wrong impression, that the algorithm already respects this constraint.
But with the ASP encoding we can search for a solution that violates the constraint by non-deterministically choosing input elements and searching for a invalid configuration (see Listing \ref{prg:val:invalid})
\begin{lstlisting}[mathescape=true,escapeinside={\#(}{\#)},basicstyle={\ttfamily\small},label=prg:val:invalid,caption={Example of constraint violation},linerange={19-19,32-32,153-154}]
1 { configuration(isA(N,elementD),0):n(N)}. 
:- not invalid_configuration(t), query(t). 
invalid_configuration(t) :- 
   N>1, element_nrofframes(A,N,t). 
\end{lstlisting}
	
If a solution can be found, we have produced a counterexample for our assumption and the algorithm is incorrect.
If no solution can be found within a given scope, there is high evidence that the algorithm is correct.
This relies on the small scope hypothesis, that most (design) errors can be found in a small scopes.
This hypothesis is also the basis for verification tools like Alloy \cite{jackson2019alloy}. 

\subsection{ UI encoding }
\label{sec:encodings:ui}

The UI encoding was inspired by an interactive configuration process using some graphical user interface (GUI).
The encoding uses these actions:

\begin{enumerate}
	\item $\mathrm{create\_element}(\mi{ELEMENTTYPE})$: create an element of the given type
 	\item $\mathrm{create\_rack}(\mi{RACKTYPE})$: create a rack of the given type
    \item $\mathrm{assign\_element\_to\_rack}(R)$: assign all the modules of the element to the given rack.
\end{enumerate}

\begin{figure}
	\includegraphics[width=\linewidth]{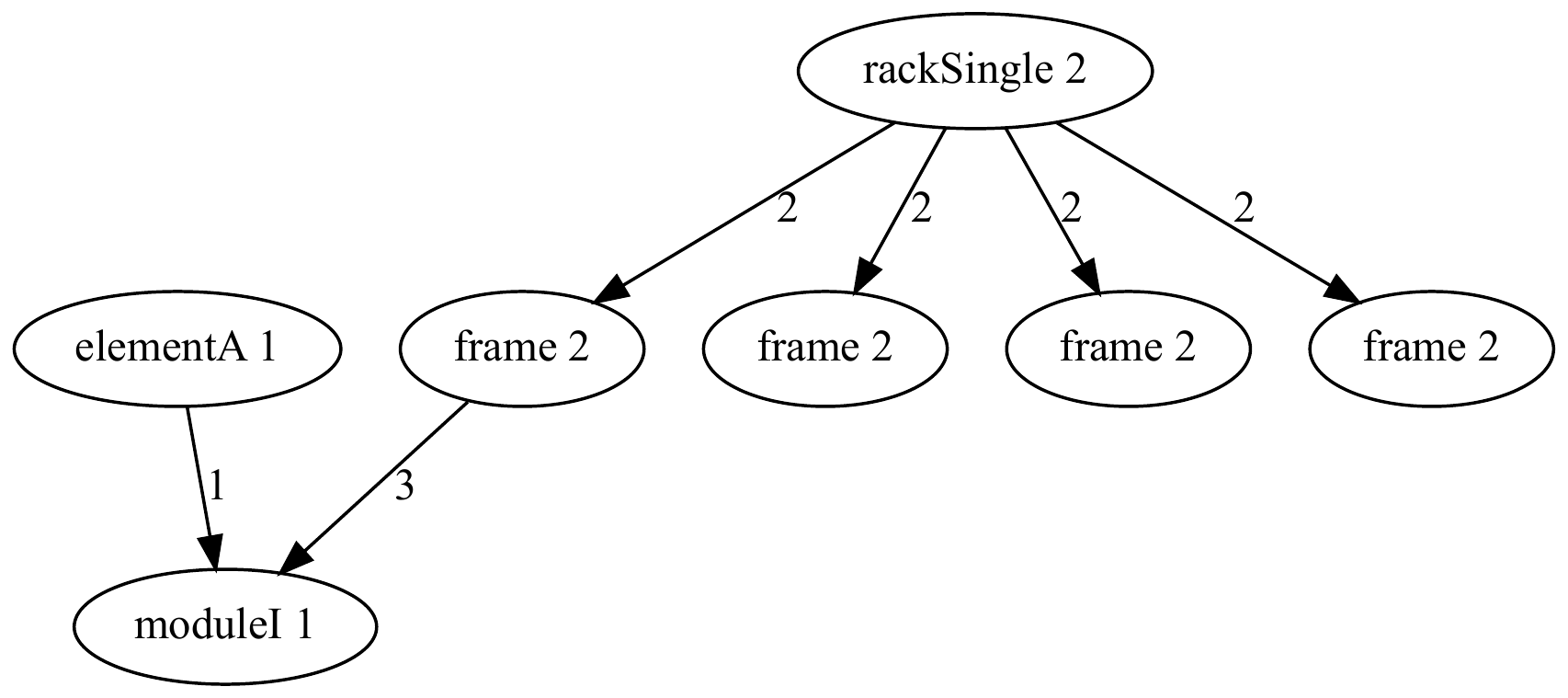}
	\caption{UI solving example}
	\label{fig:racks_ui_example}
\end{figure}

The aim of this encoding is to give the user control over the configuration process while at the same time minimizing the number of required actions.
As can be seen in \cref{fig:racks_ui_example}, the solution only needs three steps.
First the user creates an elementA, then a rackSingle, and finally the user assigns the element to the rack which indirectly assigns the module to the frames of the rack.
If the application would provide only generic user actions (create object, associate objects) like in the generic encoding, the manual configuration would require 12 steps---and this only for one element!

\paragraph{Application.}

The UI encoding enables us to reason about user actions and to answer questions such as:

\begin{itemize}
	\item Can all valid configurations be constructed with the UI actions?
 	\item Can the user create invalid configurations with the UI actions?
\end{itemize}

\section{EXPERIMENTAL RESULTS}
\label{sec:experiments}
To assess the solving performance in our various approaches, we conducted some experiments.

\subsection{Problem Instances}

Every encoding was tested on the same set of instances. 

20 racks problem instances were selected by first defining an instance-generating algorithm and then exploring instance sizes to find a set in which all systems could solve some instances under consideration within a time limit of 10 minutes. Some instances could be solved by none (or very few) of these systems.

For each racks instance $i$, with $1 \le i \le 20$, we have generated $i$ elementA, $i$ elementB, $i$ elementC, and $i$ elementD. This means, that if we consider racks instance 4, in total there are 16 element atoms defined.

In addition, in order to test the non-incremental ASP program, we had to provide an upper bound to the domain size. To this end, to each instance we add a constant $\mathrm{\#const\, domainsize}$, that needs to be large enough to allow to find a solution, but not so large to cause unnecessary grounding issues. We decided to calculate the $\mathrm{\#const\, domainsize}$ to be the domain size of the worst-case scenario configuration of each instance: in such scenario, frames and elements do not share modules, and there is a different rack for each frame. 

All solvers were configured to search for the first answer set of each problem instance.
Finding one or only a few solutions is often sufficient in industrial use cases since solving large instances can be challenging \cite{DBLP:journals/aim/FalknerFHSS16}.

\subsection{Experimental Setup}

Each of the machines used to run the experiments
was equipped with two
Intel\textsuperscript{\textregistered} Xeon\textsuperscript{\textregistered} E5-2650 v4 @ 2.20GHz CPUs with 12 cores.
Furthermore, each machine had 251 GiB of memory and ran Ubuntu 16.04.1 LTS Linux.
Scheduling of benchmarks was done with HTCondor\textsuperscript{\texttrademark}\footnote{\url{http://research.cs.wisc.edu/htcondor}} together with the ABC Benchmarking System\footnote{\url{https://github.com/credl/abcbenchmarking}} \cite{DBLP:conf/aiia/Redl16}.
Time and memory consumption were measured by \slv{pyrunlim},\!\footnote{\url{https://alviano.com/software/pyrunlim/}}
which was also used to limit time consumption to 10 minutes per instance, memory to 40 GiB and swapping to 0.
Care was taken to avoid side effects between CPUs, e.g., by requesting exclusive access to an entire machine for each benchmark from HTCondor.

The ASP system \slv{clingo}\footnote{\url{https://potassco.org/clingo/}} \cite{gebser_kaminski_kaufmann_schaub_2019} was used in version 5.4.0.

\subsection{Results}

\Cref{tab:results} shows the experimental results.
For each of the approaches described in \cref{sec:encodings}, the table contains one row with performance data.
\enquote{Generic} stands for the encoding described in \cref{sec:encodings:generic}, \enquote{Ordered} for the one in \cref{sec:encodings:ordered}, \enquote{Algorithmic} for the one in \cref{sec:encodings:algorithm}, and \enquote{UI} for \cref{sec:encodings:ui}.
\enquote{Non-incremental} refers to a non-incremental ASP encoding that we included for comparison.
The second column in \cref{tab:results} shows the number of instances solved by each approach (of the total number of 20 instances).
The third column shows the total time in minutes to process all 20 instances, where the time limit of 10 minutes was used for instances that reached the time-out.
The fourth column shows average memory consumption in GiB.

\begin{table}
	\centering
	\begin{tabular}{|c|c|c|c|}
		\hline
		\textbf{Approach} & \textbf{Solved instances} & \textbf{$\Sigma$ cputime} & \textbf{$\varnothing$ memory} \\
		\hline
		Generic & 0 & 1200 & 0.05 \\
		\hline
		Ordered & 1 & 1142 & 0.39 \\
		\hline
		Algorithmic & 20 & 132 & 0.12 \\
		\hline
		UI & 2 & 1081 & 0.45 \\
		\hline
		Non-incremental & 8 & 804 & 10.24 \\
		\hline
	\end{tabular}
	\caption{Experimental results}
	\label{tab:results}
\end{table}

\subsection{Discussion}

In this preliminary set of experiments, the algorithmic version is the only incremental encoding that outperforms the non-incremental encoding.
The non-incremental approach, however, seems to need more memory than the incremental approaches.
We assume the reason for this to be that in incremental mode, \slv{clingo} needs to ground the solution space only up to the horizon where it finds a solution, whereas in the non-incremental approach the whole solution space must be grounded upfront.

The encodings with domain-specific actions perform better than encodings with generic actions.
The generic encoding cannot even solve the smallest examples. 
This could be improved by adding some symmetry breaking constraints and general heuristics. 
On the other hand domain-specific encodings are harder to maintain and may require modifications for every new constraints. 
If the domain is too complex a deterministic algorithm might also no longer be possible.

\section{CONCLUSION}
We described how to solve product configuration problems with incremental ASP.
In doing so, we moved away from a purely declarative approach (\enquote{what}) as we also have to describe the \enquote{how} of the solving process by having to define the appropriate actions and sometimes action order to solve a particular configuration problem.
The actions of the incremental encodings can simulate the actions of a deterministic algorithm and the actions taken by an expert user during interactive configuration.

The benefit of this is that we gain more control over the solving process and that we can solve instance sizes that would be out of scope for a purely declarative approach.
The downside is the increased development effort which is similar to the effort required to develop a domain-specific solving algorithm in some imperative programming language. 
At least we stay in one knowledge representation paradigm (ASP).

For us this paper serves as a proof of concept that we can solve large configuration problems with ASP. 
To make these incremental encodings applicable in real-world projects a lot of other aspects like modularization, testing, IDE support (syntax completion for ASP, \dots) must be considered.

For future research we plan to investigate the relation between incremental and non-incremental encodings, especially if we could automatically derive configuration actions from a non-incremental encoding of the configuration problem.

Another topic is incremental repair, i.e., allowing actions that remove parts of the configuration in order to fix (upper-bound) constraint violations.

Lazy grounding \cite{DBLP:conf/inap/LeutgebW17,DBLP:conf/confws/TaupeFSW21}, as another way to ground and solve configuration problems incrementally, would also be highly interesting to investigate and compare to the multi-shot approach.

\bibliography{incremental_configuration}
\end{document}

%% file: macros.tex
\newcommand{\slv}[1]{\textsc{#1}}

\newcommand{\mi}[1]{\ensuremath{\mathit{#1}}}